\documentclass[10pt,conference,a4paper]{IEEEtran}
\usepackage{epsfig}
\usepackage{graphicx}
\usepackage{amsmath}
\usepackage{amssymb}
\usepackage{multirow}
\usepackage{tabulary}
\usepackage{tabularx}
\usepackage[tight,normalsize,sf,SF]{subfigure}
\usepackage{color}
\usepackage{soul}
\usepackage{enumitem}
\usepackage{hhline}
\usepackage{algorithm}
\usepackage{algpseudocode}
\usepackage{array}
\usepackage{flushend}
\usepackage{csquotes}
\newcolumntype{P}[1]{>{\centering\arraybackslash}p{#1}}
\renewcommand{\baselinestretch}{0.97}

\graphicspath{{./figs/}}
\DeclareGraphicsExtensions{.eps,.ps,.pdf,.png,.tiff}

\usepackage[pagebackref=true,breaklinks=true,letterpaper=true,colorlinks,bookmarks=false]{hyperref}

\ifCLASSINFOpdf
\else
\fi

\begin{document}
	
\title{Video Summarization in a Multi-View Camera Network}

\author{\IEEEauthorblockN{Rameswar Panda$^{\star}$, Abir Das$^{\dagger}$, Amit K. Roy-Chowdhury$^{\star}$}
	\IEEEauthorblockA{$^{\star}$Electrical and Computer Engineering Department, University of California, Riverside\\
		$^{\dagger}$Computer Science Department, University of Massachusetts, Lowell\\
		Email: rpand002@ucr.edu, abir.das@email.ucr.edu, amitrc@ece.ucr.edu }}

\maketitle

\begin{abstract} 
While most existing video summarization approaches aim to extract an informative summary of a single video, we propose a novel framework for summarizing multi-view videos by exploiting both intra- and inter-view content correlations in a joint embedding space. We learn the embedding by minimizing an objective function that has two terms: one due to intra-view correlations and another due to inter-view correlations across the multiple views.
The solution can be obtained directly by solving one Eigen-value problem that is linear in the number of multi-view videos.   
We then employ a sparse representative selection approach over the learned embedding space to summarize the multi-view videos. 
Experimental results on several benchmark datasets demonstrate that our proposed approach clearly outperforms the state-of-the-art.	   	 
\end{abstract}

\IEEEpeerreviewmaketitle

\section{Introduction}
\label{sec:Intro}

\begin{figure}[!t]
	\centering
	\begin{tabular}{c}
		\includegraphics[scale=0.40]{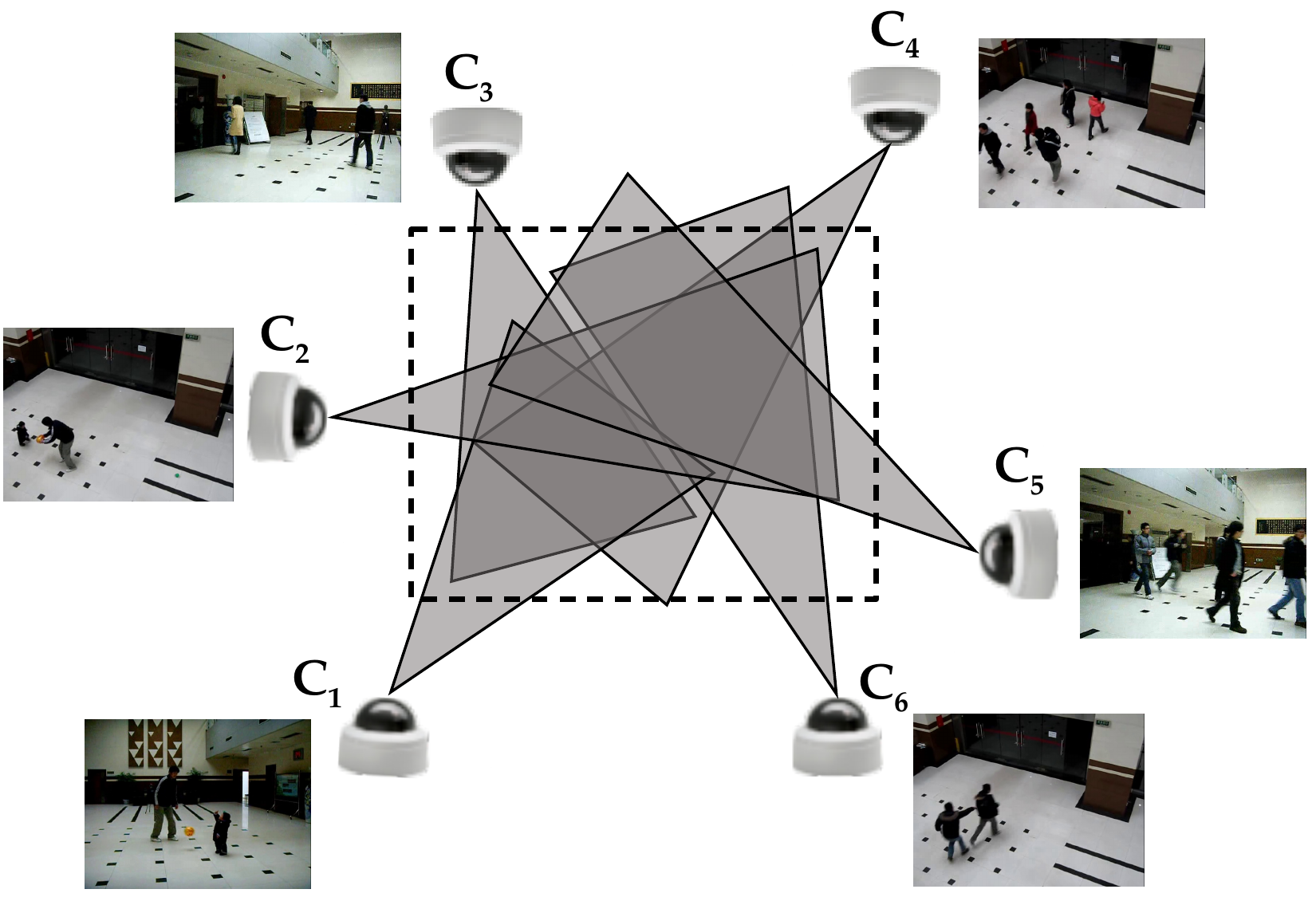}
	\end{tabular}
	\caption
	{An illustration of a multi-view camera network where six cameras $C_1$, $C_2$, \dots, $C_6$ are observing an area (black rectangle) from different viewpoints. 
	Since the views are roughly overlapping, information correlations across multiple views along with correlations in each view should be taken into account for generating a concise multi-view summary.    
	}
	\label{fig:ProblemFig} \vspace{-6mm}
\end{figure}

Network of surveillance cameras are everywhere nowadays. A major problem is to figure out how to extract useful information from the videos captured by these cameras. Fig.~\ref{fig:ProblemFig} depicts an illustrative example where a network of cameras, with both overlapping and non-overlapping fields of view (fovs) are capturing videos from a region. 
The basic question that we want to explore in such scenario is: \textit{can we get an idea of the video content without watching all the videos entirely?}

Much progress has been made in developing a variety of ways to summarize a single video, by exploring different design criteria (representativeness~\cite{Khosla2013,Ehsan2012,Eric2014,Scalable2012}, interestingness~\cite{Att2005,LucVanGool2014}) in an unsupervised manner, or developing supervised algorithms~\cite{Graumann2012,gygli2015video,gong2014diverse,Category2014}).
However, with some exceptions of~\cite{MultiviewTMM2010,MultiviewICIP2011,OnlineMultiview2015,SanjaySir2015}, summarizing multi-view videos still remains as a challenging problem because of large amount of inter-view content correlations along with intra-view correlations present in such videos.  

In this paper, we focus on the task of summarizing multi-view videos, and illustrate how a new representation that exploits multi-view correlations can effectively generate a more informative summary while comparing with the prior multi-view works. Our work builds upon the idea of subspace learning, which typically aim to obtain a latent subspace shared by multiple views by assuming that the multiple views are generated from this latent subspace. 
Specifically, our key idea is the following: by viewing two or more multi-view videos as actually being one large video, making inference about multi-view videos reduces to making inference about a single video in the latent subspace.

Our approach works as follows. First, we embed all the frames in an unified low dimensional latent space such that the locations of the frames preserve both intra- and inter-view correlations (Section~\ref{sec:Frame Embedding}). 
This is achieved by minimizing an objective function that has two terms: one due to intra-view correlations and another due to inter-view correlations across the multiple views. 
The solution can be obtained by solving an eigen-value problem that is linear in size of the multi-view videos. 
Then, we employ a sparse representative selection approach over the embedding to produce multi-view summaries (Section~\ref{sec:Summarization}). 
Specifically, we formulate the task of finding summaries as a sparse coding problem where the dictionary is constrained to have a fixed basis (dictionary to be the matrix of same data points) and the nonzero rows of sparse coefficient matrix represent the multi-view summaries.           

{\bf Contributions.} The contributions of our work can be summarized as follows.
\newline 
(1) We propose a multi-view frame embedding which is able to preserve both intra and inter-view correlations without assuming any prior correspondences/alignment between the multi-view videos. 
\newline
(2) We propose a sparse representative selection method over the learned embedding to summarize the multi-view videos, which provides \textit{scalability} in generating summaries. In particular, this allows us to generate summaries of different lengths as per the user request \textit{(analyze once, generate many)}.
\newline 
(3) The proposed method is a \textit{generalized framework} which makes sparse coding feasible in summarizing both single and multi-view videos. We demonstrate the generalizability of our framework with extensive experiments on three publicly available  multi-view datasets (11 videos) and two single view datasets (100 videos).
\newline
{\bf Related Work.} Most of the previous summarization techniques are designed for single-view videos. 
Various strategies have been studied, including clustering~\cite{VISON2012,VSUMM2011,mundur2006keyframe,furini2010stimo}, attention modeling~\cite{Att2005}, saliency based regression model~\cite{Graumann2012}, super frame segmentation~\cite{LucVanGool2014}, kernel temporal segmentation~\cite{Category2014}, crowd-sourcing~\cite{Khosla2013}, submodular maximization~\cite{gygli2015video}, and  point process~\cite{gong2014diverse}.
Interested readers can check~\cite{money2008video,Truong2007} for a more comprehensive summary.
However, they usually do not perform well for summarizing multi-view videos since they cannot exploit the large inter-view correlations.

To address the challenges encountered in a multi-view camera network, some state-of-the-art approaches use random walk over spatio-temporal shot graphs~\cite{MultiviewTMM2010} and rough sets~\cite{MultiviewICIP2011} to summarize multi-view videos. 
A very recent work in~\cite{SanjaySir2015} uses bipartite matching constrained optimum path forest clustering to solve the problem of summarizing multi-view videos. In~\cite{rameswar2016icip}, stochastic neighbor embedding with sparse coding is employed to summarize multi-view videos. 
An online method for summarization can also be found in~\cite{OnlineMultiview2015}. 
The work in~\cite{ManjunathAccv2011} and~\cite{ManjunathTON2014} also addresses a similar problem of summarization in multi-camera settings with non-overlapping field of views. 
By contrast, the approach that we describe here seeks to find summary from a multi-view network as shown in Fig.~\ref{fig:ProblemFig}. 
Moreover, our approach does not require a priori knowledge of field of view. 

\section{Multi-view Frame Embedding}
\label{sec:Frame Embedding}

{\bf Problem Statement.}
Consider a set of $K$ different videos captured from different cameras, in a $D$-dimensional space where ${X}^{(k)} = \{x_i^{(k)} \in \mathbb{R}^D ,i = 1,\cdots,N_k\}, k = 1,\cdots,K$. 
Each $x_i$ represents the feature descriptor (e.g., color, texture) of a video frame in $D$-dimensional feature space. 
As the videos are captured non-synchronously, the number of frames in each video might be different and hence there is no optimal one-to-one correspondence that can be assumed. 
We use $N_k$ to denote the number of frames in $k$-th video and $N$ to denote the total number of frames in all videos. 

Given the multi-view videos, our goal is to find an embedding for all the frames into a joint latent space while satisfying some constraints. 
Specifically, we are seeking a set of embedded coordinates ${Y}^{(k)} = \{y_i^{(k)} \in \mathbb{R}^d ,i = 1,\cdots,N_k\}, k = 1,\cdots,K$, where, $d$ $(<<D)$ is the dimensionality of the embedding space, with the following two constraints: {(1)} \textit{Intra-view correlations.} The content correlations between frames of a video should be preserved in the embedding space. {(2)} \textit{Inter-view correlations.} The frames from different videos with high feature similarity should be close to each other in the resulting embedding space as long as they do not violate the intra-view correlations present in an individual view.

{\bf Modeling Multi-view Correlations.} To achieve an embedding that preserves the above two constraints, we need to consider feature similarities between two frames in an individual video as well as across two different videos.

Inspired by the recent success of sparse representation coefficient based methods to compute data similarities in subspace clustering~\cite{elhamifar2013sparse1}, we adopt such coefficients in modeling multi-view correlations.
Our proposed approach has two nice properties: (1) the similarities computed via sparse coefficients are robust against noise and outliers since the value not only depends on the two frames, but also depends on other frames that belong to the same subspace, and (2) it simultaneously carries out the adjacency construction and similarity calculation within one step unlike kernel based methods that usually handle these tasks independently with optimal choice of several parameters. 

{\bf Intra-view Similarities.}
Intra-view similarity should reflect spatial arrangement of feature descriptors in each view. 
Based on the \textit{self-expressiveness property}~\cite{elhamifar2013sparse1} of an individual view, each frame can be sparsely represented by a small subset of frames that are highly correlated in the dataset. Mathematically, for $k$-th view, it can be represented as 
\begin{equation} 
	\begin{gathered}
		\label{eq:intra-view similarities}
		x_i^{(k)}=X^{(k)}c_i^{(k)}, \ c_{ii}^{(k)}=0,
	\end{gathered}
\end{equation} 
where $c_i^{(k)} = [c_{i1}^{(k)},c_{i2}^{(k)},...,c_{iN_k}^{(k)}]^T$, and the constraint $c_{ii}^{(k)}=0$ eliminates the trivial solution of representing a frame with itself.
The coefficient vector $c_i^{(k)}$ should have nonzero entries for a few frames that are correlated and zeros for the rest. 
However, in (\ref{eq:intra-view similarities}), the representation of $x_i$ in the dictionary $X$ is not unique in general. 
Since we are interested in efficiently finding a nontrivial sparse representation of $x_i$, we consider the tightest convex relaxation of the $l_0$ norm, i.e.,
\begin{equation}
	\begin{gathered}
		\label{eq:intra-view similarities1}
		\text{min} \ \   ||c_i^{(k)}||_1 \ \  \text{s.t.}  \ \ x_i^{(k)}=X^{(k)}c_i^{(k)}, \ c_{ii}^{(k)}=0,
	\end{gathered}
\end{equation} 
It can be rewritten in matrix form for all frames in a view as    
\begin{equation}
	\begin{gathered}
		\label{eq:intra-view similarities2}
		\text{min} \ \   ||C^{(k)}||_1 \ \  \text{s.t.}  \  X^{(k)}=X^{(k)}C^{(k)}, \ \text{diag}(C^{(k)}) = 0,
	\end{gathered}
\end{equation}
where $C^{(k)}= [c_1^{(k)},c_2^{(k)},...,c_{N_k}^{(k)}]$ is the sparse coefficient matrix whose $i$-th column corresponds to the sparse representation of the frame $x_i^{(k)}$. 
The coefficient matrix obtained from the above $l_1$ sparse optimization essentially characterize the frame correlations and thus it is natural to utilize as intra-view similarities. 
This provides an immediate choice of the intra-view similarity matrix as $C_{intra}^{(k)} = |C^{(k)}|^T$ where $i$-th row of matrix $C_{intra}^{(k)}$ represents the similarities between the $i$-th frame to all other frames in the view.

{\bf Inter-view Similarities.}
Since all cameras are focusing on roughly the same fovs from different viewpoints, all views have apparently a single underlying structure. 
Following this assumption in a multi-view setting, we find the correlated frames across two views on solving a similar $l_1$ sparse optimization like in intra-view similarities. 
Specifically, we calculate the pairwise similarity between $m$-th and $n$-th view by solving the following optimization problem:
\begin{equation}
	\begin{gathered}
		\label{eq:inter-view similarities}
		\text{min} \ \   ||C^{(m,n)}||_1 \ \  \text{s.t.}  \ \ X^{(m)}=X^{(n)}C^{(m,n)},
	\end{gathered}
\end{equation}
where $C^{(m,n)} \in \mathbb{R}^{N_n \times N_m}$is the sparse coefficient matrix whose $i$-th column corresponds to the sparse representation of the frame $x_i^{(m)}$ using the dictionary $X$.
Ideally, after solving the proposed optimization problem in (\ref{eq:inter-view similarities}), we obtain a sparse representation for a frame in $m$-th view whose nonzero elements correspond to frames from $n$-th view that belong to the same subspace. 
Finally, the inter-view similarity matrix between $m$-th and $n$-th view can be represented as $C_{inter}^{(m,n)} = |C^{(m,n)}|^T$ where $i$-th row of matrix $C_{inter}^{(m,n)}$ represent similarities between $i$-th frame of $m$-th view and all other frames in the $n$-th view. 
   
{\bf Objective Function.}
The aim of embedding is to correctly match the proximity score between two frames $x_i$ and $x_j$ to the score between corresponding embedded points $y_i$ and $y_j$ respectively.
Motivated by this observation, we reach the following objective function on the embedded points $Y$.
\begin{equation*}
\begin{gathered}
\label{eq:Completeequation}
\mathcal{F}(Y^{(1)},...,Y^{(K)}) = \sum_{k}\mathcal{F_\text{intra}}(Y^{(k)}) + \\ \sum_{m,n \atop m\neq n}\mathcal{F_\text{inter}}(Y^{(m)},Y^{(n)})
\end{gathered}
\end{equation*}
\begin{equation}
\begin{gathered}
\label{eq:Completeequation1}
= \sum_k\sum_{i,j} ||y_{i}^{(k)}-y_{j}^{(k)}||^{2}{C_{intra}^{(k)}(i,j)} + \\ \sum_{m,n \atop m\neq n}\sum_{i,j} ||y_{i}^{(m)}-y_{j}^{(n)}||^{2}{C_{inter}^{(m,n)}(i,j)}
\end{gathered}
\vspace{-2mm}
\end{equation}
where $k$, $m$ and $n = 1,\cdots,K$. $\mathcal{F_\text{intra}}(Y^{(k)})$ is the cost of preserving local correlations within $X^{(k)}$ and $\mathcal{F_\text{inter}}(Y^{(m)},Y^{(n)})$ is the cost of preserving correlations between $X^{(m)}$ and $X^{(n)}$.
The first term says that if two frames $(x_i^{(k)},x_j^{(k)})$ of a view are similar, which happens when ${C_{intra}^{(k)}(i,j)}$ is larger, their locations in the embedded space, $y_i^{(k)}$ and $y_j^{(k)}$ should be close to each other. Similarly, the second term tries to preserve the inter-view correlations by bringing embedded points $y_i^{(m)}$ and $y_i^{(n)}$ close to each other if the pairwise proximity score ${C_{inter}^{(m,n)}(i,j)}$ is high. 
The above objective function (\ref{eq:Completeequation1}) can be rewritten using one similarity matrix defined over the whole set of frames as
\begin{equation}
	\begin{gathered}
		\label{eq:Completeequation2}
		\mathcal{F}(Y) = \sum_{m,n}\sum_{i,j}||y_{i}^{(m)}-y_{j}^{(m)}||^{2}{C_{total}^{(m,n)}(i,j)} 
	\end{gathered} \vspace{-2mm}
\end{equation}
where the total similarity matrix is defined as 
\begin{equation}
	\label{eq:Total matrix}
	C_{total}^{(m,n)}(i,j) =
	\begin{cases}
		C_{intra}^{(k)}(i,j) & \text{if } \, m = n = k \\
		C_{inter}^{(m,n)}(i,j) & \text{otherwise}
	\end{cases}
\vspace{-1mm}
\end{equation}
This construction defines a $N\times N$ similarity matrix where the diagonal blocks represent the intra-view similarities and off-diagonal blocks represent inter-view similarities.
Note that an interesting fact about our total similarity matrix construction in (\ref{eq:Total matrix}) is that since each $l_1$ optimization is solved individually, a fast parallel computing strategy can be easily adopted for efficiency. 
However, the matrix in (\ref{eq:Total matrix}) is not symmetric since in $l_1$ optimization (\ref{eq:intra-view similarities1},\ref{eq:inter-view similarities}), a frame $x_i$ can be represented as a linear combination of some frames including $x_j$, but $x_i$ may not be present in the sparse representation of $x_j$. 
But, ideally, a similarity matrix should be symmetric in which frames belonging to the same subspace should be connected to each other. 
Hence, we reformulate (\ref{eq:Completeequation2}) with a symmetric similarity matrix $W=C_{total}+C_{total}^T$ as
\begin{equation}
	\begin{gathered}
		\label{eq:Completeequation3}
		\mathcal{F}(Y) = \sum_{m,n}\sum_{i,j}||y_{i}^{(m)}-y_{j}^{(m)}||^{2}{W^{(m,n)}(i,j)} 
	\end{gathered} \vspace{-2mm}
\end{equation}   
With the above symmetrization, we make sure that two frames $x_i$ and $x_j$ get connected to each other either $x_i$ and $x_j$ is in the sparse representation of the other. Furthermore, we normalize $W$ as $w_i \leftarrow w_i/||w_i||_{\infty}$ to make sure the weights in the similarity matrix are of same scale.

Given this construction, the objective function (\ref{eq:Completeequation3}) reduces to the problem of Laplacian embedding~\cite{belkin2001laplacian} of frames defined by the similarity matrix $W$. 
So, the optimization problem can be written as
\begin{equation}
	\label{eq:prox_op}
	Y^* = \underset{Y}{\operatorname{argmin}}\,\, tr\big(Y^{T}LY \big) \ \ s.t.\ \ Y^{T}DY=I,
\vspace{-2mm}
\end{equation} 
where $L$ is the laplacian matrix of $W$, $I$ is the identity matrix.
The first constraint eliminates the arbitrary scaling and avoids degenerate solutions.
Minimizing this objective function is a generalized eigenvector problem: $Ly=\lambda Dy$ and the optimal solution can be obtained by the bottom $d$ nonzero eigenvectors. 
The required embedding of the frames are given by the row vectors of $Y$.  

\section{Sparse Representative Selection}
\label{sec:Summarization}
Once the frame embedding is obtained, our next goal is to find an optimal subset of all the embedded frames, such that each frame can be described as weighted linear combination of a few of the frames from the subset. 
The subset is then referred as the informative summary of the multi-view videos. 
Therefore, our natural goal is to establish a frame level sparsity which can be induced by performing $l_1$ regularization on rows of the sparse coefficient matrix~\cite{Scalable2012,Ehsan2012}.
By introducing the row sparsity regularizer, the summarization problem can now be succinctly formulated as 
\begin{equation}
	\begin{gathered}
		\label{eq:summ equation}
		\min_{Z} \ \   ||Z||_{2,1} \ \  \text{s.t.}  \ \ Y=YZ, \ \text{diag}(Z) = 0
	\end{gathered} \vspace{-2mm}
\end{equation}
where $Z \in \mathbb{R}^{N \times N}$ is the sparse coefficient matrix and $\lVert {Z \rVert}_{2,1}\triangleq \sum_{i=1}^{N}\lVert {Z^i \rVert}_2$ is the row sparsity regularizer i.e., sum of $l_2$ norms of the rows of $Z$.
The first constraint i.e., self-expressiveness property in summarization is logical as the representatives for summary should come from the original frame set whereas the second constraint is introduced to avoid the numerically trivial solution ($Z=I$) in practice by forcing the diagonal elements to be zeros.
Minimization of (\ref{eq:summ equation}) leads to a sparse solution for $Z$ in terms of rows, \textit{i.e.}, the sparse coefficient matrix  $Z$ contains few nonzero rows which constitute the video summary. 
Notice that both of the sparse optimization in (\ref{eq:intra-view similarities2}) and (\ref{eq:summ equation}) look similar; however, the nature of sparse regularizer in both formulations are completely different. 
In (\ref{eq:intra-view similarities2}), the objective of $l_1$ regularizer is to induce element wise sparsity in a column whereas in (\ref{eq:summ equation}), the objective of $l_{2,1}$ regularizer is to induce row level sparsity in a matrix. 

The objective functions in (\ref{eq:Completeequation1}) and (\ref{eq:summ equation}) are quite general. 
One can easily notice that our framework can be extended to summarize single view videos by removing the inter-view similarities in (\ref{eq:Completeequation1}). 
Hence, our proposed embedding with sparse representative selection can summarize both single as well as multi-view videos whereas the prior sparse coding based methods~\cite{Scalable2012,Ehsan2012} can summarize only single-view videos.
Moreover, our approach is computationally efficient as the sparse coding is done in lower-dimensional space and at the same time, it preserves the locality and correlations among the original frames which has a great impact on the summarization output.         
         
\section{Solving the Sparse Optimization Problems}
\label{sec:Optimization1}
We solve all the sparse optimization problems using an Alternating Direction Method of Multipliers (ADMM) framework~\cite{boyd2011distributed}.
Due to space limitation, we only present the optimization procedure to solve (\ref{eq:summ equation}).
However, the same procedure can be easily extended to solve other sparse optimizations (\ref{eq:intra-view similarities2},~\ref{eq:inter-view similarities}). 
Using Lagrange multipliers, the optimization problem (\ref{eq:summ equation}) can be written as
\vspace{-2mm}
\begin{equation}
	\begin{gathered}
		\label{eq:summ equation1}
		\min_{Z}\ \lVert {Z\rVert}_{2,1} + \dfrac{\lambda}{2}\lVert {Y - YZ \rVert}^2_{F} \\
		s.t.\ \ diag(Z)=0
	\end{gathered} \vspace{-1mm}
\end{equation}
where $\lambda$ is the regularization parameter that balances the weight of the two terms. 
To facilitate the optimization, we consider an equivalent form of (\ref{eq:summ equation1}) by introducing an auxiliary variable $A$:
\vspace{-2mm}
\begin{equation}
	\begin{gathered}
		\label{eq:summ equation2}
		\min_{Z,A}\ \lVert {Z\rVert}_{2,1} + \dfrac{\lambda}{2}\lVert {Y - YA \rVert}^2_{F} \\
		s.t.\ \ A=Z,\ \   diag(Z)=0
	\end{gathered} \vspace{-1mm}
\end{equation}  
ADMM tries to solve (\ref{eq:summ equation2}) by iteratively updating $A$ and $Z$ shown in Algo.~\ref{algo:Sparse Dictionary Selection}, where the shrinkage-thresholding operator $S_{\mu}(z)$ acting on each row of the given matrix is defined as
\begin{equation}
	\begin{gathered}
		\label{eq:summ equation3}
		S_{\mu}(z)= \max  \Big\{||z||_2-\mu,0\Big\}\frac{z}{{\lVert z \rVert}_2}.
	\end{gathered} \vspace{-1.5mm}
\end{equation} 

\vspace{-3mm}
\begin{algorithm} 
	\caption{An ADMM solver for (\ref{eq:summ equation1})}\label{algo:Sparse Dictionary Selection}
	\begin{algorithmic}
		\State {\bf Input:} Embedded feature matrix $Y$ 
		\State {\bf Initialization:} Initialize $A,Z,B$ to zero and $\lambda, \rho >0$
		\While{\textit{not converged}} \\
		\ \ $A \leftarrow (\lambda Y^{T}Y+\rho I)^{-1}(\lambda Y^{T}Y+\rho Z-B)$;\\
		\ \ $A \leftarrow A-\text{diag}(\text{diag}(A))$; \\
		\ \ $Z \leftarrow S_{\frac{1}{\rho}}(A+B/\rho)$ (row-wise);\\
		\ \ $Z \leftarrow Z-\text{diag}(\text{diag}(Z))$; \\
		\ \ $B \leftarrow B+\rho(A-Z)$;
		\EndWhile
		\State {\bf Output:} Sparse coefficient matrix $Z$.
	\end{algorithmic}
\end{algorithm} 
\vspace{-2mm}          

Since the problem (\ref{eq:summ equation1}) is convex, Algo.~\ref{algo:Sparse Dictionary Selection} is guaranteed
to converge by the existing ADMM theory~\cite{glowinski1989augmented}.

 
{\bf Remark 1.} We do not require to compute $(\lambda Y^{T}Y+\rho I)^{-1}$ in each iteration for updating $A$. More specifically, it is unchanged during iterations of Algo.~\ref{algo:Sparse Dictionary Selection}. Thus, one can pre-compute the required Cholesky factorizations to avoid redundant computations for efficiently solving those linear systems. 
\section{Experiments}
\label{sec:Experiments}
{\bf Datasets.} We conduct rigorous experiments using five publicly available datasets: (i) Office dataset captured with 4 stably-held web cameras in an indoor environment~\cite{MultiviewTMM2010}, (ii) Campus dataset taken with 4 handheld ordinary video cameras in an outdoor scene~\cite{MultiviewTMM2010}, (iii) Lobby dataset captured with 3 cameras in a large lobby area~\cite{MultiviewTMM2010}, (iv) Open Video Project (OV) dataset of 50 videos~\cite{VISON2012} and (v) YouTube dataset of 50 videos~\cite{VISON2012}.
These datasets are extremely diverse: while the first three datasets consists of multi-view videos with overall 360 degree coverage of the scene, the last two datasets contains single view videos of several genres. 

{\bf Features.} We utilize Pycaffe with the \enquote{BVLC CaffeNet} pretrained model~\cite{jia2014caffe} to extract a 4096-dim CNN feature vector (i.e. the top layer hidden unit activations of the network) for each video frame. 
We use deep features, as they are the state-of-the-art visual features and have shown best performance on various recognition tasks. 

{\bf Performance Measures.} We compare all the approaches using three quantitative measures, including Precision, Recall and F-measure~\cite{MultiviewTMM2010}.
For all these metrics, the higher value indicates better summarization quality.

{\bf Other Details.} The regularization parameters $\lambda$ is taken as $\lambda_{0}/\gamma$ where $\gamma > 1$ and $\lambda_{0}$ is analytically computed from the data~\cite{Ehsan2012}.
In Algo.~\ref{algo:Sparse Dictionary Selection}, the stop criteria is defined as following:
\vspace{-1.5mm}
\begin{equation}
\begin{gathered}
\label{eq:stop}
||A^{(t)}-Z^{(t)}||_{\infty} \leq \epsilon  \ or \ t \geq 2000 
\end{gathered} \vspace{-2mm}
\end{equation}
where $t$ is the iteration number and $\epsilon$ is set to $10^{-7}$ throughout the experiments.  

\begin{table*} [t]
	\centering
	\caption{Performance comparison with several baselines including both single and multi-view methods applied on the three multi-view datasets. \textbf{P}: Precision in percentage, \textbf{R}: Recall in percentage and $\bf{F}$: {F}-measure. Ours perform the best.
	} 	
	\vspace{-2mm}
	\label{tab:Multi-view Comparison Table}
	\begin{tabulary}{1.1\linewidth}{|p{30mm}|P{10mm}|P{10mm}|P{10mm}|P{10mm}|P{10mm}|P{10mm}|P{10mm}|P{10mm}|P{10mm}|}
		\hline
		
		& \multicolumn{3}{c|}{\textbf{Office}} &\multicolumn{3}{c|}{\textbf{Campus}}&\multicolumn{3}{c|}{\textbf{Lobby}}\\
		\cline{2-10}
		
		\textbf{Methods} &\textbf{P}& \textbf{R} & $\bf{F}$ &\textbf{P}& \textbf{R} & $\bf{F}$ &\textbf{P}& \textbf{R} & $\bf{F}$ \\
		\hhline{|=|=|=|=|=|=|=|=|=|=|}
		ConcateAttention~\cite{Att2005}	& 100	& 38	& 55.07 & 56	& 48	& 51.86 & 31	& 95& 81.98 \\
		ConcateSparse~\cite{Ehsan2012}	& 100	& 54 & 66.99 & 59	&45  & 50.93 &93 	&65  & 76.69	\\ 
		AttentionConcate~\cite{Att2005}	& 100	& 46 & 63.01 & 40	& 28 & 32.66 & 100	& 70 & 82.21	\\
		SparseConcate~\cite{Ehsan2012}	& 94	& 58  & 71.34 	& 58	&52  &  54.49	&88 &70  & 77.87\\
		RandomWalk~\cite{MultiviewTMM2010} 	& 100	& 61 & 76.19 & 70	& 55 & 61.56 & 100	& 77 & 86.81	\\ 
		RoughSets~\cite{MultiviewICIP2011} 	& 100	& 61  & 76.19& 69 & 57 & 62.14 & 97 & 74&84.17\\
		BipartiteOPF~\cite{SanjaySir2015} & 100	& 69 & 81.79 & 75	& 69 & 71.82 & 100 & 79 & 88.26	\\ 
		\textbf{Ours}	& \textbf{100}	& \textbf{73} & \textbf{84.48} & \textbf{84}	& \textbf{69} & \textbf{75.42} & \textbf{100}	& \textbf{79} & \textbf{88.26}\\ \hline 
		
	\end{tabulary} \vspace{-2mm}
\end{table*} 
    
{\bf \underline{Multi-view Video Summarization:}} This experiment aims at evaluating our proposed framework in summarizing multi-view videos compared to the state-of-the-art including both methods for single view and multi-view video summarization.

{\bf Compared Methods.} We compare our approach with total of seven existing approaches including four baseline methods (ConcateAttention~\cite{Att2005}, ConcateSparse~\cite{Ehsan2012}, AttentionConcate~\cite{Att2005}, SparseConcate~\cite{Ehsan2012}) that use single-view summarization approach over multi-view datasets to generate summary and three methods (RandomWalk~\cite{MultiviewTMM2010}, RoughSets~\cite{MultiviewICIP2011}, Bipartite-OPF~\cite{SanjaySir2015}) which are specifically designed for multi-view video summarization.
Note that the first two baselines (ConcateAttention, ConcateSparse) concatenate all the views into a single video and then applie a summarization approach, whereas in the other two baselines (AttentionConcate, SparseConcate), an approach is first applied to each view and then the resulting summaries are combined along the time line to form a single multi-view summary.
~\cite{Scalable2012} also uses the same objective function as in~\cite{Ehsan2012} for summarizing consumer videos. 
The only difference lies in the algorithm used to solve the objective function (proximal vs ADMM). Hence, we compared only with~\cite{Ehsan2012}.      
The purpose of comparing single view methods is to show that techniques that attempt to find informative summary from single-view videos usually do not produce an optimal set of representatives while summarizing multi-view videos.
We employ the ground truth of important events reported in~\cite{MultiviewTMM2010} for a fair comparison. 
In our approach, an event is taken to be correctly detected if we get a representative frame from the set of ground truth frames between the start and end of the event. 
\begin{figure*}[!t]
	\centering
	{
		\includegraphics[width=0.75\linewidth]{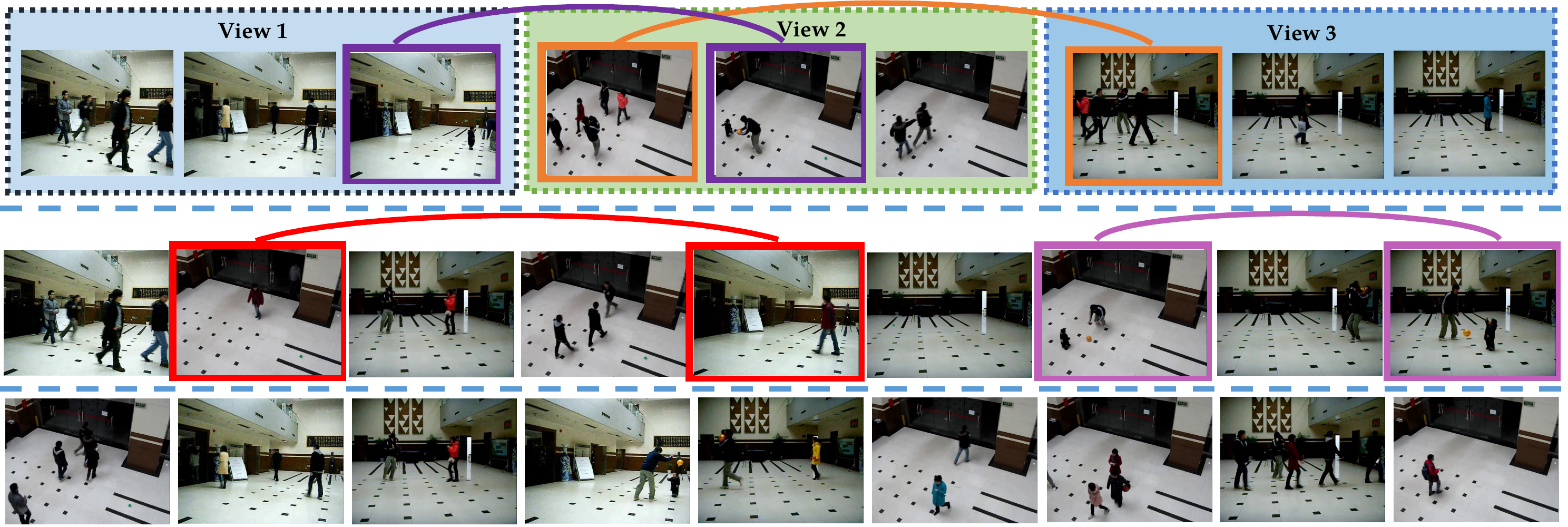}}
	\vspace{-2mm}
	\caption
	{Some summarized events for the \textit{Lobby} dataset. 
		Top row: summary produced by SparseConcate~\cite{Ehsan2012}, Middle row: summary produced by ConcateSparse~\cite{Ehsan2012}, and Bottom row: summary produced by our approach. It is clearly evident from both top and middle rows that both of the single-view baselines produce a lot of redundant events in summarizing multi-view videos, however, our approach (bottom row) produces meaningful representatives by exploiting the content correlations via an embedding.
		Redundant events are marked with same color borders. 
		Note that, although the frames with same color border look somewhat visually distinct, they essentially represent same events as per the ground truth in~\cite{MultiviewTMM2010}.         
	}
	\label{fig:View-board}\vspace{-5mm}
\end{figure*} 
\begin{figure} 
	\centering
	{
		\includegraphics[width=1\linewidth]{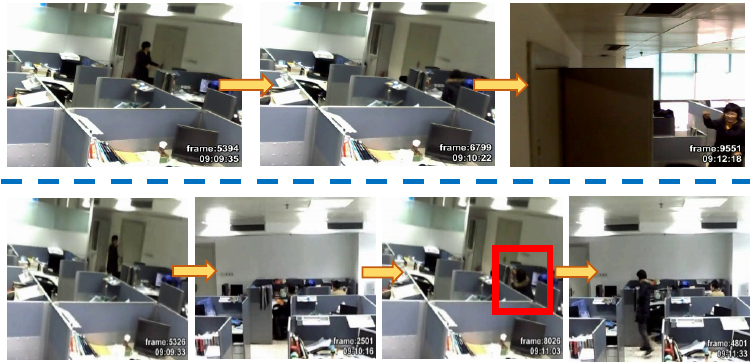}}
	\vspace{-6mm}
	\caption
	{Sequence of events detected related to activities of a member $(A_0)$ inside the Office dataset. Top row: Summary produced by method~\cite{MultiviewTMM2010}, and Bottom row: Summary produced by our approach. 
	The event of looking for a thick book to read (as per the ground truth in~\cite{MultiviewTMM2010}) is missing in the summary produced by method~\cite{MultiviewTMM2010} where as it is correctly detected by our approach (3rd frame: bottom row). This indicates our method captures video semantics in more informative way compared to~\cite{MultiviewTMM2010}.         
	}
	\label{fig:Event Detection} 
	\vspace{-6.5mm}
\end{figure}   
 
{\bf Results.} Table.~\ref{tab:Multi-view Comparison Table} show the summarization results on all three multi-view datasets.
The analysis of the results for both Office and Lobby dataset are quite interesting in two aspects. 
First, our approach produces summaries with same precision as RandowWalk for both of the datasets. 
However, the improvement in recall value indicates the ability of our method in keeping more important information in the summary compared to RandomWalk.
One such illustrative example for the Office dataset is presented in Fig.~\ref{fig:Event Detection}.
Second, our performance is similar to the recently published baseline BipartiteOPF for Lobby dataset but we improved around 5\% in terms recall and 3\% in terms of F-measure for the Office dataset.

Notice that for all methods, including ours, performance on Campus dataset is not that good as compared to other two datasets. 
This is obvious since the Campus dataset contains many trivial events as it was captured in an outdoor environment, thus making the summarization more difficult. 
Nevertheless, for this challenging dataset, F-measure of our approach is about 15\% better than that of RandomWalk and 5\% better than that of BipartiteOPF.     
Overall, on all datasets, our approach outperforms all the baselines in terms of F-measure.
This corroborates the fact that sparse representative selection coupled with multi-view frame embedding produces better summaries in contrast to the state-of-the-art methods.        

Furthermore, while comparing with several mono-view summarization approaches (ConcateAttention, ConcateSparse, AttentionConcate, SparseConcate), Table.~\ref{tab:Multi-view Comparison Table} reveals that summaries produced using these methods contain a lot of redundancies (simultaneous presence of most of the events) since they fail to exploit the complicated inter-view frame correlations present in multi-view videos.
However, our proposed framework significantly outperforms these methods in terms of precision, recall and F-measure due to its ability to model multi-view correlations.
Limited to the space, we only present a part of the summarized events for the Lobby dataset as illustrated in Fig.~\ref{fig:View-board}.

{\bf \underline{Generalization to Single-view Video Summarization:}} The objective of this experiment is to validate the generalizabilty of our framework in summarizing single view videos along with multi-view videos. In particular, the basic question that we want to explore in this experiment is: \textit{does the learned embedding also help in summarizing single-view videos?} 

{\bf Compared Methods.} We contrast our approach with several baselines covering a wide variety of single-view methods as follows: (1) DT~\cite{mundur2006keyframe} that model a video using a delaunay triangulation to extract the key frames, (2) STIMO~\cite{furini2010stimo}: uses a fast clustering algorithm with advanced user customization, (3) VSUMM~\cite{VSUMM2011}: uses an improved k-means algorithm clustering and then the centroids are deemed as key frames, (4) VISON~\cite{VISON2012}: this method extract key frames based on local maximum in the frame similarity curve, (5) a sparse coding approach that does not consider frame correlations (Sparse)~\cite{Scalable2012,Ehsan2012}. 
We follow the standard procedure in~\cite{VSUMM2011} to obtain the mean performance measures on comparing with all user-created summaries.

{\bf Results.} Table.~\ref{tab:Single-view Comparison Table} shows results on both OV and YouTube datasets. 
Without surprise, Sparse~\cite{Ehsan2012} performs better as compared to other clustering based methods~\cite{mundur2006keyframe,furini2010stimo,VSUMM2011,VISON2012} since it selects key frames based on how representative a particular frame is in the reconstruction of the original video.    
However, our method performs even better compared to Sparse. 
We believe the improvement can be attributed to frame embedding that exploits frame correlations in sparse representative selection. 
\begin{table} 
	\centering
	\caption{Performance of various single view video summarization methods on both OV and YouTube datasets.}
	\vspace{-2mm}
	\label{tab:Single-view Comparison Table}
	\begin{tabulary}{1.1\linewidth}{|p{17mm}|P{6mm}|P{6mm}|P{6mm}|P{6mm}|P{6mm}|P{6mm}|}
		\hline
		
		& \multicolumn{3}{c|}{\textbf{OV}} &\multicolumn{3}{c|}{\textbf{YouTube}}\\
		\cline{2-7}
		
		\textbf{Methods} &\textbf{P}& \textbf{R} & $\bf{F}$ &\textbf{P}& \textbf{R} & $\bf{F}$ \\
		\hhline{|=|=|=|=|=|=|=|}
		DT~\cite{mundur2006keyframe}	& 67.7	& 53.2	& 57.6 & 40.7& 42.8	& 42.3 \\
		STIMO~\cite{furini2010stimo}	& 60.3 & 72.2 & 63.4 & 46.2 & 43.1 & 45.6	\\ 
		VSUMM~\cite{VSUMM2011}	& 70.6	& 75.8 & 70.3	& 58.3	& 57.6 & 56.8 \\
		VISON~\cite{VISON2012}	& 70.1	& 82.0 & 75.5	& 50.1	& 51.5 & 49.2 \\
		Sparse~\cite{Ehsan2012} 	& 79.5 	& 83.4 & 78.2 &65.7 & 63.8 & 61.2	\\ 
		\textbf{Ours} 	& \textbf{81.6}	& \textbf{84.8} & \textbf{80.4} & \textbf{67.0}	& \textbf{66.5} & \textbf{64.6} \\
		\hline 
		
	\end{tabulary} \vspace{-5mm}
\end{table} 

{\bf \underline{Scalability in Generating Summaries:}}
The aim of this experiment is to demonstrate the scalability of our approach in generating summaries of different length based on the user constraints without any further analysis of the input videos. 
Such property makes sense in surveillance systems as one user may want to see only 5 most important events of the day whereas at the same time, another user may want to see only 2 most important events that occurred in the whole day.  

\begin{figure} [h]
	\centering
	{
		\includegraphics[width=1\linewidth]{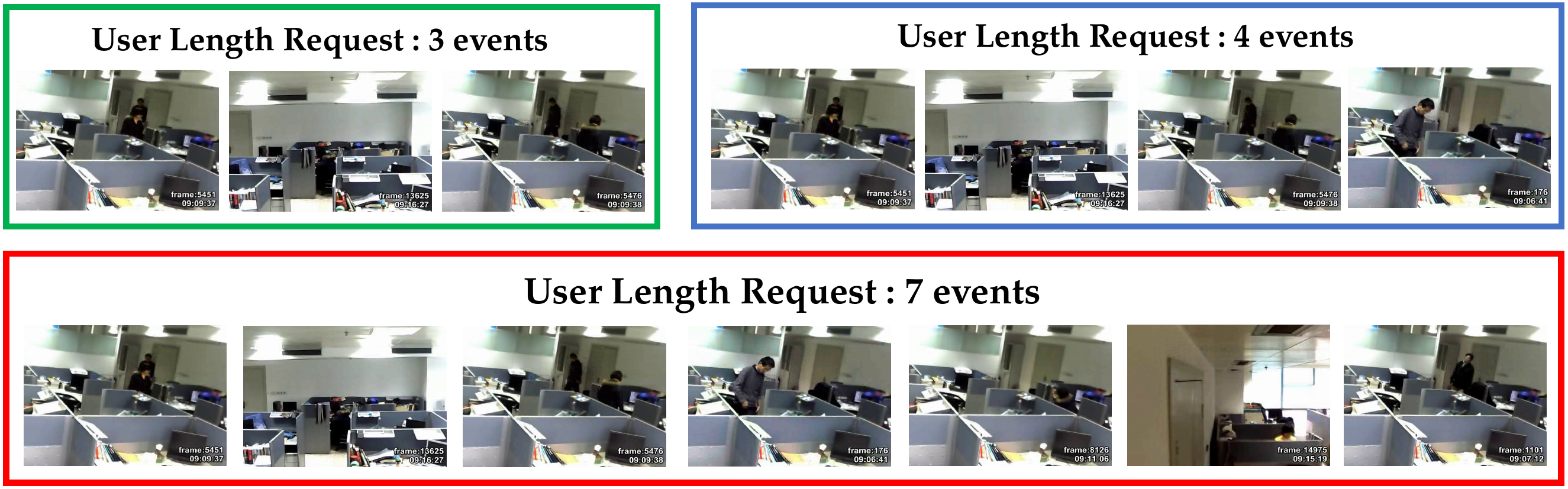}}
	\vspace{-6mm}
	\caption
	{The figure shows an illustrative example of scalability in generating summaries of different length based on the user constraints for the Office dataset. 
		Each event is represented by a key frame and are arranged according to the $l_2$ norms of corresponding non-zero rows of $Z$. 
	}             
	\label{fig:Scalability} \vspace{-1mm}
\end{figure}

{\bf Results.} Apart from indicating the representatives for
the summary, the non-zero rows of $Z$ also provides information about the relative importance of the representatives for describing the whole videos. 
A higher ranking representative frame takes part in the reconstruction of many frames in the multi-view videos as compared to a lower ranked frame. 
This provides scalability to our approach as the ranked list can be used as a scalable representation to provide summaries of different lengths (\textit{analyze once, generate many}). 
Fig.~\ref{fig:Scalability} shows the generated summaries of length 3, 4 and 7 most important events (as determined by the method described above) for the Office dataset. 
\vspace{-3mm}
\section{Conclusions}
\label{sec:Conclusions}
In this paper, we present a novel framework for summarizing multi-view videos in a camera network by exploiting the content correlations via an joint embedding. The embedding formulation introduced encodes both intra and inter-view correlations in a unified latent subspace.
We then employ a sparse coding method over the embedding that provides scalability in generating the summaries. 
We show the effectiveness of our framework through rigorous experimentation on five datasets.

{\bf Acknowledgments:} This work was partially supported by NSF grant IIS-1316934. 

{\small
	\bibliographystyle{ieee}
	\bibliography{ICPR_2016}
}

\end{document}